\documentclass[conference, compsoc]{IEEEtran}

\usepackage{graphicx,amsmath, amssymb}
\usepackage{latexsym}

\begin{document}

\title{Learning Taxonomy for Text Segmentation by Formal Concept Analysis}

\author{\IEEEauthorblockN{Mihaiela Lupea}
        \IEEEauthorblockA{
         "Babe\c s-Bolyai" University\\ Cluj-Napoca, Romania}
\and
        \IEEEauthorblockN{Doina Tatar}
        \IEEEauthorblockA{"Babe\c s-Bolyai" University\\ Cluj-Napoca, Romania}
\and
        \IEEEauthorblockN{ Zsuzsana Marian}
        \IEEEauthorblockA{"Babe\c s-Bolyai" University\\ Cluj-Napoca, Romania}}

\date{}

\maketitle
\begin{abstract}
In this paper the problems of deriving a taxonomy from a text and concept-oriented text segmentation are approached. Formal Concept Analysis (FCA) method is applied to solve both of these linguistic problems. The proposed segmentation method offers a conceptual view for text segmentation, using a context-driven clustering of sentences. The  Concept-oriented Clustering Segmentation algorithm {\it (COCS)} is based on k-means linear clustering of the sentences. Experimental results obtained using {\it COCS} algorithm are presented.

\end{abstract}

\section{ Introduction}
Formal Concept Analysis (FCA) studies how objects can be hierarchically grouped together
when their common attributes are studied in a given context.
Linguists often characterize datasets using distinct features, such as semantic components or syntactical and grammatical markers, which can easily be interpreted using FCA.
  However, linguists argue that formal concepts are quite different from cognitive processes relating to natural language \cite{UtaPriss}. This is why current FCA applications in linguistics focus more on formal structures than on cognitive linguistic phenomena.

 Eventually, in the linguistic  domain FCA
applications provide a very suitable  alternative to statistical
methods.

In this paper we address  the problem of deriving a taxonomy from a text for text segmentation by concept-driven clustering.
This conceptual view of segmentation is useful when different users have quite
different needs with regard to way of segmentation.

 The needed knowledge in our {\it  Concept-oriented  Clustering Segmentation algorithm COCS}  is only  the taxonomy derived from text. It is used the k-means algorithm for a linear clustering of the sentences.

The paper is structured as follows:
Section 2 introduces  the basic notions of ontologies and FCA. Section 3 surveys the related work in taxonomies extraction from a text  and in text segmentation.  Section 4 introduces {\it CLTE} (concept lattice-taxonomy extraction) algorithm and {\it COCS} algorithm  for text segmentation.
  In Section 5  experimental results obtained using {\it COCS} algorithm are presented. We finish the paper with conclusions and future work directions in Section 6.

 \section{Abstract Ontologies and FCA}

  Following \cite{GanWille}, an ontology is {\it a formal specification of a shared conceptualization of a domain of interest to a group of users}. {\it Formal} implies that the ontology should be machine readable, and {\it shared} implies it is accepted by a group or community.

 {\bf Definition 1.} An abstract ontology $O$ is a model represented by:
   $$ O= (C,H,R,A) $$
   where:
    \begin{itemize}

      \item{$C$ is a set of concepts (concept identifiers);}

      \item{ $H$ is a taxonomic relation ( IS-A) between concepts, $H \subseteq C \times C$ , that means it is a partial and transitive order on $C$;}

      \item{$R$ is a set of non-taxonomic relations, $R \subseteq C \times C$;}


      \item{ $A$ is a set of logical axioms (or inference rules).}

    \end{itemize}


Mostly approaches focus on the first two elements of an ontology $C $ and $H$, which form the "core ontology" while the researches on the sets $R$ and  $A$ are least addressed.

 The above definition doesn't make a distinction between a concept and its lexical expression. Completing $O$ with a lexicon could be addressed  the problems of synonymy (a set of lexical expressions represents the same concept) and that of the polysemy (a lexical  expression represents a set of concepts).

 In the particular case of learning  a taxonomy  from a text
we will present the method used by  \cite{cimiano1}  and our proposed version.

\subsection{A short survey of Formal Concept Analysis (FCA)}

    FCA has been introduced by B. Ganter and R. Wille in 1982 (for a textbook see \cite{GanWille}).
     During the last years, FCA has grown into an international research community with applications
          in many different domains as artificial intelligence, linguistics, software engineering, medicine, etc..
          Formal concepts in FCA can be seen as a mathematical formalization of what has been called the theory of concepts, which states that a concept is formally defined via its features \cite{UtaPriss}.
     From a philosophical point of view, a concept is a unit consisting from two parts: the {\it extension} (the set of objects belonging to this concept) and the {\it intension} (the set of attributes valid for all these objects). The frame for defining a set of concepts
    is the so called {\it Formal Context }.

 {\bf Definition 2.} A {\it Formal Context } is a triple:
      $$ K=(G,M, I)$$
where $G$ is the set of objects, $M$ is the set of attributes, and $I$ is a binary relation between $G$ and $M$ ($I\subseteq G \times M$), representing the incidence relation. The pair $(g,m)\in I$ is read as "the object $g$ has the attribute $m$". \\

Usually a {\it Formal Context } is given by an incidence
       matrix, where a star "$*$"  on the line of $g$ and  the column of $m$ means that
       the object $g$ has the attribute $m$.

        For a set $A \subseteq G$, the set of all attributes shared by the objects from $A$, called  the "derivative" of $A$ and denoted by $A'$  is defined as:
         $$ A'= \{m\in M \mid \forall g \in A, (g,m)\in I\} $$

       Dually, for  a set $B \subseteq M$, the set of all objects which share  the attributes  from   $B$, called the "derivative" of $B$  and denoted by $B'$ is defined as:
         $$ B'= \{g\in G \mid \forall m \in B, (g,m)\in I\} $$

{\bf Definition 3.} A {\it Formal Concept} of the {\it Formal Context } $K=(G,M,I)$ is a pair $(A,B)$, with $A\subseteq G$, $ B\subseteq M$ and satisfying the relations:
   $$ A'=B \;\;\;\;\; and \;\;\;\;\; B'=A$$

    The set $ A$ is called the {\it extent} of the {\it Formal Concept} $(A,B)$ and the set $ B$ is called the {\it intent} of the same {\it Formal Concept}.

  Between the {\it Formal Concepts} the relation $\leq$ of subconcept-superconcept is defined as
  below:
     $$(A_1,B_1) \leq (A_2,B_2)\;\; if\;\; and\;\; only\;\; if\;\;  A_1\subseteq A_2 $$
      or equivalently
$$ (A_1,B_1) \leq (A_2,B_2)\;\; if\;\; and\;\; only\;\; if\;\;B_2 \subseteq B_1. $$

  The set of all {\it Formal Concepts } of a {\it Formal Context}, $K$, together with the order relation $\leq$ forms o complete lattice called the {\it Concept lattice}, and denoted $B(K)$. The top (the last element of the {\it Concept lattice}) is  $1_{B(K)}$ and the  bottom (the first element of the {\it Concept lattice}) is   $0_{B(K)}$.

   Each node $X$ of the lattice is characterized by a set of objects $A$ and a set of attribute $B$. The set  $A$ is formed by all the objects situated on paths which begin with $X$ (including $X$) and end on the bottom of lattice, and the set $B$  is formed by all the attributes  situated on paths which begin with top and end on $X$ (including $X$). Moreover,
 $A'=B$ and $B'=A$ and thus the  node labeled  by the pair $(A,B)$  represents a
 {\it Formal Concept}.



{\bf Remarks}:1. Each object and attribute  is introduced at a single node.
2. The objects situated lower (higher) in the lattice have more (less) attributes.
3. The attributes situated lower (higher) in the lattice are shared by less (more) objects.

  Rules for simplifying the {\it Concept lattice} are applied when they are not clarified and have the objects and the attributes reducible:

  {\bf Definition 4.}
     A {\it Concept lattice}  is {\it clarified} if no two of its objects have the equal intents, and no two of its attribute have the equal extents. These properties could be observed from the incidence matrix of the {\it Formal Context}.

 {\bf Definition 5.}
     An attribute $m$ of a clarified {\it Formal Context} is {\it reducible} if there is a set $S \subseteq M$ of attributes such that $\{m\}'=S$, otherwise it is {\it irreducible}. {\it Reducible} objects are defined dually.

{\bf Remark}: If $m$ is reducible, it can be deleted from the {\it Formal Context} (dually for a reducible object).

    Reading from this {\it Concept lattice}  the labels which introduce  {\it attributes} and transforming the obtained lattice in tree such that all the inheritances  between attributes are  kept, a taxonomic hierarchy is obtained (see Sections 3 and  4).

\section{Related work}

\subsection{ Automatic learning of a taxonomy from a text: Cimiano's approach}

        The most well known work in automatic learning of a taxonomy from a text is given by the Karlsruhe's team \cite{cimiano1},\cite{cimiano2}.
      Let us present the example introduced in \cite{cimiano1} for obtaining a taxonomy from a text on the tourism domain using FCA.

       The {\it Formal Context} is obtained selecting as $M$ the set of transitive verbs from a text and as $G$ the set of nouns playing the role of (direct) complement for the verbs from $M$.

      For the selected domain:

 {\small\it M=\{bookable, rentable, driveable, rideable,
 joinable\}},

 {\small\it G=\{apartment, car, excursion, motor-bike, trip, hotel\}}

\noindent  and the relation $I $ is given by the incidence matrix  (Table 1).



\begin{table}

\begin{center}

\begin{tabular}{*{6}{|c}|}
 \hline
\multicolumn{1}{|c|}{} &
\multicolumn{1}{|c|}{book.} & \multicolumn{1}{|c|}{rent.} &
\multicolumn{1}{|c|}{drive.} & \multicolumn{1}{|c|}{ride.}& \multicolumn{1}{|c|}{join.} \\
\hline
\hline $apartment$ &  $*$ & $*$  & $ - $ & $- $ & $- $ \\

\hline $car$ & $*$ & $*$ & $*$ & $ -$& $- $  \\

\hline $motor - bike$ & $*$ & $*$ & $*$ & $
*$ & $- $  \\

\hline $excursion$ & $*$ & $-$ & $ -$ & $ -$ & $* $ \\

\hline $trip$ & $*$ & $-$ & $ -$ & $ -$ & $* $ \\

\hline $hotel$ & $*$ & $-$ & $ -$ & $ -$ & $- $ \\

\hline

\end{tabular}
\end{center}

\caption{The incidence matrix for  tourism example }

\label{t1}

\end{table}


  \begin{table}

\begin{center}

\begin{footnotesize}

\begin{tabular}{*{3}{|c}|}
 \hline
\multicolumn{1}{|c|}{Extent of  concept} &
\multicolumn{1}{|c|}{Intent of  concept} & \multicolumn{1}{|c|}{Concept} \\

\hline
\hline $\{apartment,car,$ &  $$ & $$   \\
 $motor-bike,trip,$ &  $ \{bookable\}$ & $C_1 $   \\
 $excursion,hotel\}$ &  $ $ & $ $   \\
\hline $\{apartment,car,$ & $\{bookable,rentable\}$ & $C_2$   \\
$motor-bike\}$ & $$ & $$   \\

\hline $\{ car,motor-bike\}$ & $\{bookable,rentable,$ & $C_3$ \\
$$ & $driveable \}$ & $$ \\
\hline $\{motor-bike\}$ & $\{bookable,rentable,$ & $C_4$  \\
$$ & $driveable,rideable \}$ & $$  \\

\hline $\{excursion, trip\}$ & $\{bookable,joinable \}$ & $C_5$  \\

\hline $$ & $\{bookable,rentable,$ & $$  \\
$\Phi$ & $driveable,rideable,$ & $C_6$  \\
$$ & $joinable \}$ & $$  \\

\hline

\end{tabular}

\end{footnotesize}

\end{center}

\caption{The  {\it Formal Concepts} for tourism example }

\label{t2}

\end{table}

According to the method for obtaining the {\it Concept Lattice} (\cite{GanWille}), the set of all {\it Formal Concepts} are represented in  Table 2.
Applying the definition of the {\it subconcept} relation, the following {\it Concept lattice}, is obtained:

\begin{verbatim}
              C1 bookable/(hotel)
             /  \
            /    \
           /      \
rentable/ C2       C5 joinable/
(apartment)|       /({excursion,trip})
           |      /
driveable/ C3    /
(car)      |    /
           |   /
rideable/ C4  /
(motor-bike)\/
            C6
\end{verbatim}

Let us remark that the lattice is not {\it clarified} because the set of objects:$\{$ {\it excursion, trip}$\}$ have the same intent: $\{${\it bookable, joinable}$\}$. This is the reason in the node $C_5$ the "object" label is formed by the set $\{${\it bookable, joinable}$\}$.

{\bf Example 1.} Consider the concept $C_1$=$(A_1,B_1)$ from the previous lattice.
 Here the extent $A_1$ is formed by all the objects situated on paths starting with $C_1$. $A_1 =G=\{apartment,car, motor-bike,excursion,trip,hotel\}$. The intent is  $B_1= \{bookable\}$.
The relations $A_1'=B_1$ and $B_1'=A_1$ are verified.

For the concept $C_2$=$(A_2,B_2)$ the extent is $A_2$=$\{apartment, car, motor - bike\}$
and the intent is $B_2= \{bookable, rentable\}$. Again, the relations $A_2'=B_2$ and $B_2'=A_2$ are verified. The {\it Concept lattice} relation $C_2\leq C_1$ is valid, because  $A_2 \subseteq A_1$ (and, equivalently,  $B_1 \subseteq B_2$ ).

   From the {\it Concept lattice} of the tourism example the following taxonomy is obtained
   \cite{cimiano1} :
\begin{verbatim}
            bookable
             / |    \
            /  |     \
      joinable |      rentable
          /|   |      /    \
         / |   |     /      \
 excursion |   | apartament  \
           |   |        driveable
           |   |           /\
         trip  |          /  \
               |        car  rideable
               |               |
               |               |
            hotel         motor-bike
\end{verbatim}

{\bf Remark}: In this kind of taxonomy the name of verbs could be replaced by the  name of corresponding noun: for example {\it joinable} could be replaced by {\it join} or
{\it driveable} by {\it vehicle} to improve the readability  of the taxonomy.


As we already have mentioned above, in \cite{cimiano1}
the {\it Formal Context} is obtained selecting as $M$ the set of transitive verbs from a text and as $G$ the set of nouns playing the role of (direct) complement for the verbs from $M$ (subcategorized by the transitive verbs in $M$).
 It is possible to obtain pairs of object/attribute which are in a false position of
 complement/verb and to lose other pairs, when the corpus is not large enough. To improve this probability Cimiano  clustered the nouns and the verbs using a vectorial model and finally he  considered clusters of nouns as objects and clusters of verbs as attributes, instead of nouns and verbs.

    To obtain the vectors he considered the conditional probability  $P(n \mid v)$, where  $P(a \mid b) =\frac{f(a,b)}{f(b)} $.  Here $f(n,v)$ represents the frequency of occurrences of the noun $n$ as a complement of the verb $v$.
     An improved   value of $P(n \mid v)$  is obtained by realizing before a noun and verb clustering \cite{cimiano2}. For this goal he calculated for
                each noun $n$  and verb $v$ the vectors:
           $ V_n=(P(n\mid v_1), \cdots, P(n\mid v_l))$ and $ V_v=(P(n_1\mid v), \cdots, P(n_k\mid v))$
and defined the similarity between nouns and between verbs as:
             $ sim(n_1,n_2)=cosine(V(n_1),V(n_2))$ and $ sim(v_1,v_2)=cosine(V(v_1),V(v_2))$.

 At each step he recalculated all $P(n|v)$ where the clustered nouns $n$ are considered together, and then  the clustered verbs $v$ are considered together.
      He alternated noun clustering and verb clustering until $P(n|v)$ is above some threshold. The obtained clusters of nouns and verbs represent objects and attribute, respectively. The incidence relation between $n$ and $v$ means the occurrence of an element from the cluster of $n$ as a complement of an element of the cluster of $v$.

  Cimiano also proposed (\cite{cimiano2})  relation: verbs (as objects) and nouns-subject (as attributes) and showed that using both these dependencies leads to better results.

\subsection{Related works in Segmentation}

  A discourse segment consists of a sequence of sentences that
  display local coherence. Text segmentation is the automatic
  identification of boundaries between segments. The need for
  discourse segmentation derives from its applicability in many fields as for example:

  \begin{itemize}

   \item {\it Information Retrieval} (IR). Many authors, like  \cite{Hearst}
    and \cite{Reynar}, showed
   that   segmenting into distinct
   topics is useful as IR needs to find relevant portions of text
   that match with a given query;

   \item {\it Anaphora resolution} (AR). Mining the text only in some
   segments for finding the antecedents for some referential
   expressions could improve the quality of AR (\cite{Mitkov});

   \item {\it Text summarization}. Segmentation as a pre-processing step in automatic
   summarization (as in this paper) could improve the quality of summaries
   \cite{bog}.

  \end{itemize}

  While the need for segmentation of discourse is almost
  universally agreed upon, there is no consensus on how the
  segmentation could be accomplished \cite{ja}. However,  a
   classification of the main directions of segmentation  is as follows:

\begin{itemize}

\item Topical text segmentation relies on  finding the sentences
that will be borderlines (topic's shifts) in the discourse. The
applied method is usually the calculation of similarity which
measures proximity between sequences of  sentences or clauses
(\cite{Hearst});

\item  Lexical chains segmentation methods rely on  lexical
chains  which display the cohesion  that arises from semantic
relationships  between words, relationships derived from WordNet
or Roget's Thesaurus (\cite{Regina}, \cite{Kept2009});

\item Referential analysis segmentation methods act in the way
that  if a referring expression is used that requires an
antecedent situated in a previous sentence, then all sentences
between the antecedent and referring expression are considered to be in
the same segment (\cite{Mitkov});

\item Earlier discourse  segmentation methods are
 Rhetorical Structure Theory (\cite{MaTo}) or Hobbs's coherence relations
  \cite{Hobbs} based on
 cue phrases  (for example {\it anyway} is an end of
 a digression in attentional stack method \cite{ja} and {\it
 because} is a causal relation in RST theory) and on a large taxonomy of different
 relations that can  hold between sentences and segments.

\end{itemize}

Another classification of segmentation methods relies on the
structure type of the output. In  linear segmentation the
discourse is divided into a linear sequence of adjacent segments
( \cite{Hearst} or this paper) while in
hierarchical segmentation there are hierarchically organized sets of
segments, as for example attentional/intentional structures of
Grosz and Sidner (\cite{Grosz}), rhetorical trees in RST theory
(\cite{MaTo}) or attentional stacks in \cite{ja}.  Recently a new method of linear segmentation has been  proposed  in \cite{tkst} which uses a kind of
{\it complementing} set of formal concepts in concept lattice  of a given formal context.

 A final  classification of segmentation methods is into cohesion based methods
 (as for example lexical chains) and coherence based methods (as
  in RST theory and Hobbs's coherence relations theory).

\section{This paper proposal}

\subsection{Obtaining the Concept Lattice and the Concept Hierarchy from a text}
FCA is used to build the Concept Lattice and then to extract the Concept Hierarchy from a text using as attributes the transitive verbs and as objects the corresponding nouns with the role of direct complements from the studied corpus.
We propose Concept Lattice - Taxonomy Extraction {\it (CLTE)} algorithm
which introduces specific rules for deriving the taxonomy as a quasi-tree from the Concept Lattice.\\

{\bf Concept Lattice - Taxonomy Extraction algorithm} ({\it CLTE}):

{\it Input}: {\it Text} - a text document.

{\it Output}: $K$-the formal context, $L$-the concept lattice,

\hspace {0.9cm} $T$- the taxonomy based on the concept lattice.

{\it Step1}: {\it Text-Pos} = Pos-tagging({\it Text}).

{\it Step2}: {\it Pairs} = $\{$(verb, noun-direct-complement)$\}$;

\hspace{1.8cm}= extract-pairs({\it Text-Pos}).

{\it Step3}: {\it Pairs-lemma}=lemmatize-verbs-nouns({\it Pairs}).

{\it Step4}: $M$ = frequent-verbs({\it Pairs-lemma});

\hspace {0.9cm} $G$ = frequent-nouns({\it Pairs-lemma}).

{\it Step5}: Build the formal context: $K=(G,M,I)$

\hspace{0.9cm} where $(n,v)\in I,\;\;if\;\; (v,n)\in$ {\it Pairs-lemma}.

{\it Step6}: Build the concept lattice $L$=$B(K))$.

{\it Step7}: Build the taxonomy $T$, represented as a

\hspace{1cm}quasi-tree, based on the concept lattice $L$.

{\bf Remarks}:
\begin{itemize}
\item
The POS annotation is enough and no parsing is needed for the initial text corpus. Rules for determining the dependency {\it verb - noun as a direct complement} must be used.

\item
Generally the taxonomy, derived from a concept lattice, cannot be represented as a tree like in Cimiano's example, but using a special data structure, called a {\it quasi-tree} (a node may have more parents and two internal nodes may have the same label), $T=(X,E)$, with the following properties:
   \begin{itemize}
   \item{ $X=G \bigcup M$ and $E$, the set of edges, is obtained from de subconcept relation of the {\it Concept lattice} according to special rules.}
   \item{The most general concept (the top of the lattice) is the root of the quasi-tree}.
   \item{The leaves of the quasi-tree $T$  are labeled with nouns (objects) from $G$ and the internal nodes are labeled with verbs (attributes) from $M$}.
   \item Let $C^{o,a} \rightarrow C^{o',a'}$ be an edge in the {\it Concept lattice}, where the node $C^{o,a}$ introduces the object $o$ and the attribute $a$. There are 16 cases ($a,o,a',o'$ can be equal or not equal with $\emptyset$), some of them impossible cases. The most used rules for adding nodes and edges in the taxonomy, represented as a quasi-tree, are the following:
   \begin {itemize}
   \item
      if $a \neq\emptyset, o = \emptyset, a'\neq \emptyset$ then $(a,a')\in E$;
   \item
      if $a=\emptyset, o\neq \emptyset, a'=\emptyset$ then $(a,a')\in E$;
   \item
      if $a\neq \emptyset, o\neq \emptyset, a'=o'=\emptyset$ then $(a,o)\in E$, $o$ is a leaf node;
   \item
      if $a\neq \emptyset, o\neq \emptyset, a'\neq\emptyset$ then $(a,a')\in E$, $(a,o)\in E$, $o$ is a leaf node;
    \item
      if $a\neq \emptyset, a'=\emptyset, o'\neq\emptyset$ then $(a,o')\in E$, $o'$ is a leaf node;
      \item
      if $a=\emptyset, o=\emptyset, a'=\emptyset, o'\neq\emptyset$ then $(a,o')\in E$, $o'$ is a leaf node;
        \item
      if $a\neq\emptyset, o=a'=o'=\emptyset$ then $(a,a)\in E$;
   \end{itemize}

   \end{itemize}
\item
   A path from the root to a leaf node provides a hierarchy regarding the concept terms (verbs and nouns) on that path.
\end{itemize}

\subsection{Concept-oriented segmentation by clustering}

The process of segmentation is seen as an objective method, which provides one clearly defined result.  However, different users have quite different needs with regard to a segmentation  because  they view the same text from completely different, subjective, perspective.
  Segmenting  a text must be associated with an explanation of why
  a given set of segments  is produced.
   All these could be realized by viewing  the process of segmentation as a clustering process of the sentences of a text \cite{tkst}.

  When  the cluster $Cl=\{ S_{i_1},\cdots, S_{i_m}\}$ is one   of the set of obtained clusters, and $i_1 \leq i_2 \cdots \leq i_m$ , then the linear segmentation is:
  $[S_1,S_{i_1-1}][S_{i_1},S_{i_2}], \cdots, [S_{i_{m-1}},S_{i_m}], [S_{i_m},S_n]  $. The concept terms which are "specific"  to this cluster $Cl$  (concept terms  specific to the center of cluster $Cl$) explain the reason of the segmentation.

  Let us remark that usually clustering texts means selecting of the most important (by frequency) words (terms) as features  of clustering (\cite{cimiano2}). In our method we  choose as
   words  the transitive verbs and complement nouns which  form the concepts in the FCA approach ({\cite{cimiano1}). In what  follows we refer to these words (terms) as {\it concept terms}, namely {\it concept attribute terms}, $M$,  and   {\it concept object terms}, $G$.

   A sentence is represented as a vector of concept terms: an entry of each vector specifies the frequency that a concept term occurs in the text, including the frequency of subconcept terms.

The following algorithm   is an improvement of an own algorithm introduced  in \cite{tkst}.\\

{\bf Concept-oriented  Clustering  Segmentation  algorithm} {\it COCS}:

{\bf Input:}
   $Text=\{S_1, \cdots, S_n\}$ of $n$ sentences,

\hspace{0.9cm}- the output of the CLTE algorithm:

\hspace{0.9cm}$K$- the formal context, $L$- the concept lattice;

\hspace{0.9cm}$T$- the taxonomy based on $L$.

{\bf Output:}
  Different segmentations of the text $Text$,\hspace{2cm} according to different sets of  concepts.
\begin{itemize}
\item
{\bf Step1:} Calculate the frequency $f(i,t)$ of the concept term $t \in G \cup M$ in the sentence $S_i$.
\item
{\bf Step2:} {Calculate the total frequency of the concept term $t$ in the  sentence $S_i$ as
      $Total_S(i,t)=f(i,t) + \sum_{t' \;is\; a \;direct\; descendent\; of \;t \;in\;the \;taxonomy} f(i,t') $.}

\item
{\bf Step3:}
   {Calculate the total  frequency of $t$ for all sentences as $Total(t)=
  \sum_{i=1}^{n} Total_S(i,t)$.}

\item
{\bf Step4:} {Choose the first $m=\frac{1}{2}|G\cup M|$ best supported concept terms: $t_1,\cdots,t_m$ (which maximize $Total(t)$).}

\item {\bf Step4:}
   {Represent each sentence $S_i$ by a {\it m-concept term vector}:

     $V(i)=(Total_S(i,t_1),\cdots,Total_S(i,t_m))$}.

\item {\bf Step5:}
   {Apply a linear clustering of the set of sentences $Text=\{S_1, \cdots, S_n\}$, using {\it K-means algorithm}, where
  $sim(S_i,S_j)=cosine(V(i),V(j))$ }
\end{itemize}

A cluster corresponds to a segmentation  as above. The {\it concept terms} specific for this cluster explain the "view" of segmentation and help the user to understand the differences between clustering (segmentation) results. \\
\\
 The used clustering method is {\bf K-means} which we survey below.

{\bf K-means algorithm}\cite{Manning}:

{\bf Input:}  $Text=\{S_1, \cdots, S_n\}$ of $n$ sentences,
 the corresponding vectors $\{V(1), \cdots, V(n)\}$ obtained at Step4 of {\it COCS} algorithm.

{\bf Output:} The set of clusters $C=\{C_1,C_2,...,C_k\}$

Begin

\hspace{0.2cm}     {\it Select k initial centroids}:

\hspace{0.8cm} $\{\stackrel{\rightarrow}{f_1},\stackrel{\rightarrow}{f_2},...,\stackrel{\rightarrow}{f_k}\}\subset \{V(1), \cdots, V(n)\}$

\hspace{0.2cm} While {\it the stopping criterion is not true} Do

\hspace{0.7cm}For j=1 to $k$ Do

\hspace{0.7cm} $C_j=\{V(i)|\forall \stackrel{\rightarrow}{f_l},
d(V(i),\stackrel{\rightarrow}{f_j})\leq d(V(i),\stackrel{\rightarrow}{f_l}), $

\hspace{0.7cm} $ d(V(i),V(j))=\frac{1}{cosine(V(i),V(j))}\}$

\hspace{0.7cm}End-For

\hspace{0.7cm}For j=1 to $k$ Do

\hspace{0.8cm}$\stackrel{\rightarrow}{f_j}=\frac{\sum_{\stackrel{\rightarrow}{x}\in C_j}\stackrel{\rightarrow}{x}}{|C_j|}$

\hspace{0.7cm}End-For

\hspace{0.2cm}End-While

End-algorithm\\

The {\it K-means algorithm} begins with a set of initial cluster centers, selected such that they are as least similar as possible. At each while-iteration, each vector is assigned to the cluster whose center is closest and then the {\it centroids} of the modified clusters are recomputed as a {\it mean} of its members. The distance between two vectors is computed as the inverse of the similarity of the vectors. The stopping criterion can be the condition that the diameters of all clusters are smaller than a threshold value or that there are no changes in $C$ from the previous iteration. A diameter of a cluster is the distance between the least similar elements in the cluster.


\section{Experimental results}
The algorithms proposed in the previous sections were implemented and tested on texts from different domains as art, music, law.

\begin{figure*}
	\centering
		\includegraphics[scale=0.8]{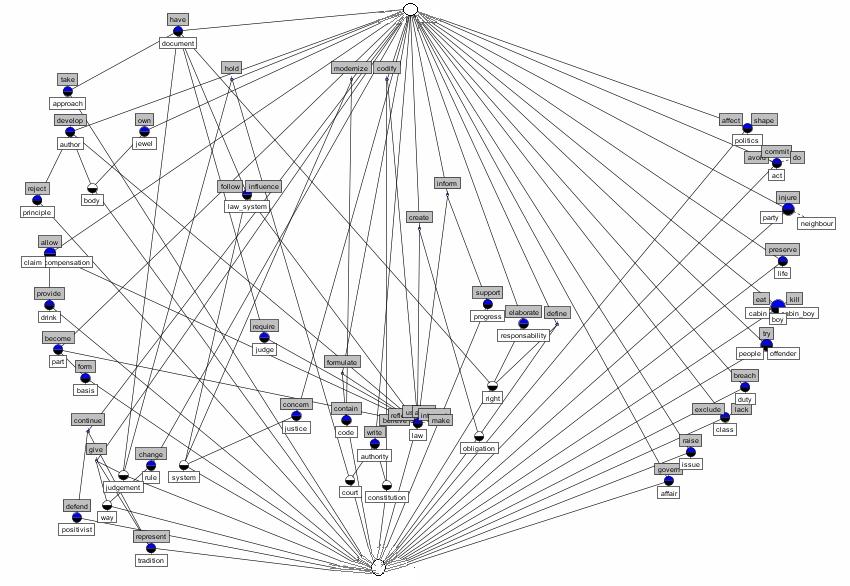}
	\caption{Concept Lattice for a corpus in the law domain}
	\label{fig:context_l}
\end{figure*}

Considerations for implementation:
\begin{itemize}
\item
For POS-tagging and lemmatization of verbs and nouns we have used Online CST tools which incorporate a {\it tokenizer, name recognizer, Brill-POS-tagger} (an error-driven transformation-based tagger), {\it lemmatiser, NP recognizer} and other tools (http://conexp.sourceforge.net/index.html).

\item
The pairs (transitive verb, noun as a direct complement) were obtained using our specific rules for determining this type of dependency.
\item
The most frequent verbs and nouns were choosed such that they appear twice in the set of selected pairs.
\item
The construction from the concept lattice of the quasi-tree representing the taxonomy of the concept terms is based on the rules proposed in Subsection 4.1.
\item
The implementation of {\it COCS}-algorithm follows the described above steps.

\end{itemize}

As experimental results we describe an example of a text, consisting of 320 sentences, from the law domain. An extract of 30 sentences occurs in the Figure 2.  The Concept lattice is computed with the {\it CLTE} algorithm and visualized in  Figure 1, using ConExp.  This  is a software tool aimed for handling the tasks involved in the study of lattice theory, mainly formal concepts. (More information is available at http://conexp.sourceforge.net/index.html.)

\begin{figure}[h!]
{
\footnotesize

\vspace{0.1cm}
\hrule
\vspace{0.1cm}
{\bf Text}:
\vspace{0.1cm}
\hrule
\vspace{0.1cm}
1.Law is a system of rules, usually enforced through a set of institutions.
2.It shapes politics, economics and society in numerous ways and serves as a primary social mediator of relations between people.
3.Contract law regulates everything from buying a bus ticket to trading on derivatives markets.
4.Property law defines rights and obligations related to the transfer and title of personal (often referred to as chattel) and real property.
5.Trust law applies to assets held for investment and financial security, while tort law allows claims for compensation if a person's rights or property are harmed.
6.If the harm is criminalized in a statute, criminal law offers means by which the state can prosecute the perpetrator.
7.Constitutional law provides a framework for the creation of law, the protection of human rights and the election of political representatives.
8.Administrative law is used to review the decisions of government agencies, while international law governs affairs between sovereign nation states in activities ranging from trade to environmental regulation or military action.
9.Writing in 350 BC, the Greek philosopher Aristotle declared, "The rule of law is better than the rule of any individual."
10.Legal systems elaborate rights and responsibilities in a variety of ways.
11.A general distinction can be made between civil law jurisdictions, which codify their laws, and common law systems, where judge made law is not consolidated.
12.In some countries, religion informs the law.
13.Law provides a rich source of scholarly inquiry, into legal history, philosophy, economic analysis or sociology.
14.Law also raises important and complex issues concerning equality, fairness and justice.
15."In its majestic equality", said the author Anatole France in 1894, "the law forbids rich and poor alike to sleep under bridges, beg in the streets and steal loaves of bread."
16.In a typical democracy, the central institutions for interpreting and creating law are the three main branches of government, namely an impartial judiciary, a democratic legislature, and an accountable executive.
17.To implement and enforce the law and provide services to the public, a government's bureaucracy, the military and police are vital.
18.While all these organs of the state are creatures created and bound by law, an independent legal profession and a vibrant civil society inform and support their progress.
19.Constitutional and administrative law govern the affairs of the state.
20.Constitutional law concerns both the relationships between the executive, legislature and judiciary and the human rights or civil liberties of individuals against the state.
21.Most jurisdictions, like the United States and France, have a single codified constitution, with a Bill of Rights.
22.A few, like the United Kingdom, have no such document.
23.A "constitution" is simply those laws which constitute the body politic, from statute, case law and convention.
24.A case named Entick v Carrington illustrates a constitutional principle deriving from the common law.
25.Mr Entick's house was searched and ransacked by Sheriff Carrington.
26.When Mr Entick complained in court, Sheriff Carrington argued that a warrant from a Government minister, the Earl of Halifax, was valid authority.
27.However, there was no written statutory provision or court authority.
28.The great end, for which men entered into society, was to secure their property.
29.That right is preserved sacred and incommunicable in all instances, where it has not been taken away or abridged by some public law for the good of the whole ...
30.If no excuse can be found or produced, the silence of the books is an authority against the defendant, and the plaintiff must have judgment.
\vspace{0.1cm}
\hrule
}
\caption{ Law domain Text }
\label{f1}
\end{figure}

The taxonomy is too complex to be depicted, but we present some paths in the corresponding quasi-tree representing hierarchies of concept terms:
\begin{itemize}
\item
$inform\rightarrow support \rightarrow progress$
\item
$continue \rightarrow represent \rightarrow tradition$
\item
$have \rightarrow influence \rightarrow law-system$
\item
$ codify \rightarrow make \rightarrow law$
\item
$ develop \rightarrow reject \rightarrow principle$
\end{itemize}

The {\it COCS} algorithm was applied only to the first 102 sentences of the initial text. At Step2, the frequency of a concept term {\it t} in a sentence is obtained as the sum of its own frequency and the frequencies of the direct descendents of {\it t} in the taxonomy.
For example: $Total_S(14, concern)=f(14,concern)+f(14,justice)+f(14,system)=1+1+0=2$,

There are 21 terms (rpresenting the value of {\it m} in Step4 of COCS algorithm): $\{${\small \it concern, have, kill, law, own, offenders, include, write, boy, condone, preserve, eat, hold, do, create, make, govern, provide, buy, shape, jewel}$\}$ used as features for clustering.
After the clustering process 4 clusters were obtained.

The cluster C1=$\{S_8,S_{19},S_{27},S_{31},S_{37},S_{40},S_{60},S_{63}\}$ is characterized by the concept terms:$\{${\small \it have, offenders, write, condone, do, govern}$\}$, meaning that these terms appear in the sentences of the cluster.\\
The corresponding linear segmentation of the text is: $\left[S_1,S_7\right]$,
$\left[S_8,S_{18}\right]$, $\left[S_{19},S_{26}\right]$, $\left[S_{27},S_{30}\right]$, $\left[S_{31},S_{36}\right]$, $\left[S_{37},S_{39},\right]$, $\left[S_{40},S_{59}\right]$, $\left[S_{60},S_{62}\right]$, $\left[S_{63},S_{102}\right]$.

The cluster C2=$\{S_{3},S_{14},S_{20},S_{53},S_{54},S_{68},S_{71},$
$S_{74},S_{84}\}$ is characterized by the concept terms: $\{${\small \it concern, preserve, buy, shape, jewel}$\}$ and provides the segmentation:
$\left[S_1,S_2\right]$,
 $\left[S_{3},S_{13}\right]$, $\left[S_{14},S_{19}\right]$, $\left[S_{20},S_{52}\right]$, $\left[S_{53},S_{53},\right]$, $\left[S_{54},S_{67}\right]$, $\left[S_{68},S_{70}\right]$, $\left[S_{71},S_{73}\right]$,
$\left[S_{74},S_{83}\right]$, $\left[S_{84},S_{102}\right]$.

\section{Conclusions and further work}

          In  this paper we applied the FCA theory to obtain a taxonomy (algorithm {\it CLTE}) for  concept-oriented segmentation   of a text.
       The  {\it COCS} algorithm introduced in this paper approaches the process of segmentation as a clustering process of the sentences  of a text, using the  taxonomy learned  from a text. Each cluster provides a segmentation, explained by the concept terms specific for this cluster.

       As further work we propose to improve the taxonomy learned from a text considering also as pairs of attribute-object: (verb at the passive, corresponding noun with the role of subject). More experiments with texts from different domains are needed in order to evaluate our approach.

\end{document}